# PathoScribe: Transforming Pathology Data into a Living Library with a Unified LLM-Driven Framework for Semantic Retrieval and Clinical Integration

Abdul Rehman Akbar[1, *], Samuel Wales-McGrath[1], Alejadro Levya[1, ‡], Lina Gokhale[1, ‡], Rajendra Singh[2, 3], Wei Chen[1], Anil Parwani[1], Muhammad Khalid Khan Niazi[1]

[1] Department of Pathology, College of Medicine, The Ohio State University Wexner Medical Center, Columbus, OH, USA

[2] Department of Pathology and Laboratory Medicine, Perelman School of Medicine, The University of Pennsylvania, Philadelphia, PA, USA

[3] PathPresenter

[‡] The authors contributed equally to this work.

[*] Correspondence: Abdul.Akbar@osumc.edu (A.R.A.)

## Abstract

Pathology underpins modern diagnosis and cancer care, yet its most valuable asset, the accumulated experience encoded in millions of narrative reports, remains largely inaccessible. Although institutions are rapidly digitizing pathology workflows, storing data without effective mechanisms for retrieval and reasoning risks transforming archives into a passive data repository, where institutional knowledge exists but cannot meaningfully inform patient care. True progress requires not only digitization, but the ability for pathologists to interrogate prior similar cases in real time while evaluating a new diagnostic dilemma.

We present PathoScribe, a unified retrieval-augmented large language model (LLM) framework designed to transform static pathology archives into a searchable, reasoning-enabled living library. PathoScribe enables natural language case exploration, automated cohort construction, clinical question answering, immunohistochemistry (IHC) panel recommendation, and prompt-controlled report transformation within a single architecture.

Evaluated on 70,000 multi-institutional surgical pathology reports, PathoScribe achieved perfect Recall@10 for natural language case retrieval and demonstrated high-quality retrieval-grounded reasoning (mean reviewer score 4.56/5). Critically, the system operationalized automated cohort construction from free-text eligibility criteria, assembling research-ready cohorts in minutes (mean 9.2 minutes) with 91.3% agreement to human reviewers and no eligible cases incorrectly excluded, representing orders-of-magnitude reductions in time and cost compared to traditional manual chart review.

By unifying retrieval and reasoning, PathoScribe enables pathologists to move from isolated case interpretation toward data-informed decision-making grounded in institutional precedent. This work establishes a scalable foundation for converting digital pathology archives from passive storage systems into active clinical intelligence platforms.

## Introduction

The practice of pathology is fundamental to modern medicine, underpinning diagnosis, prognosis, and treatment planning for a vast array of diseases, most notably cancer [1, 2]. The primary source of this critical information resides in pathology reports, which are rich, narrative documents created by pathologists [3-6]. However, the inherent unstructured nature of these reports, combined with the sheer volume of data generated by traditional pathology workflows, presents a significant barrier to large-scale clinical research and efficient decision support at the point of care [5, 7, 8]. Extracting structured, actionable insights from millions of free-text reports for tasks such as building patient cohorts for research or commercialization, identifying relevant historical cases, or standardizing clinical recommendations remains a labor-intensive and error-prone process [9, 10].

Clinicians and researchers frequently face the challenge of locating relevant historical cases within institutional archives, a task that can require manual review of hundreds and thousands of reports and may take days to weeks for complex queries [11]. This bottleneck becomes even more pronounced when seeking institutional precedents for rare diagnoses or constructing research cohorts with strict inclusion and exclusion criteria.

Despite decades of progress in digital pathology, most platforms remain predominantly image-centric, lacking the robust natural language understanding (NLU) capabilities required to unlock the clinical value embedded within free-text pathology reports [12-26]. This absence of scalable, semantically aware tools for information retrieval and cohort construction represents a critical gap in current informatics infrastructure—one that directly hinders diagnostic efficiency and research productivity. By accumulating petabytes of unstructured data without semantic accessibility, we risk transforming our collective medical library into a "digital dump," where institutional knowledge is effectively lost to a silent fire of obscurity rather than utilized for patient care. From a patient perspective, true progress is not measured by the sheer volume of data archived, but by a clinician's ability to reason through institutional history at the point of care.

Recent advances in natural language processing (NLP) and large language models (LLMs) have transformed the ability to interpret complex biomedical text, enabling impressive performance on tasks such as information extraction, clinical question answering, and decision support [27-30]. Domain-adapted models trained on pathology corpora have further demonstrated promise in structured data extraction and report classification [31]. However, existing applications remain narrowly focused on individual tasks such as report structuring or diagnostic coding, failing to address the broader spectrum of clinical workflows that pathologists navigate daily [32]. Moreover, most computational pathology efforts have concentrated on whole slide image (WSI) analysis, leaving the textual component of pathology practice relatively underexplored despite its central role in clinical communication and knowledge synthesis.

To address this gap, we introduce PathoScribe, a unified pathology intelligence framework that leverages LLMs to enable natural language interaction with institutional pathology archives (Figure 1). We conceptualize this capability as a living library of pathology, an archive that is not merely stored, but computationally searchable, semantically interpretable, and capable of informing real-time clinical reasoning. Transforming static repositories into such living systems requires tightly integrating large-scale retrieval with contextual understanding. Accordingly, the system combines retrieval-augmented generation (RAG) architecture with domain-adapted language models to support multiple clinical workflows from a single infrastructure [33, 34]. Unlike traditional keyword-based search systems or structured query languages that require specialized syntax, our framework accepts free-text queries and complex clinical criteria expressed in natural language, democratizing access to institutional knowledge for clinicians regardless of technical expertise. From a diagnostic perspective, this framework empowers pathologists to perform real-time comparative analysis against rare or complex historical precedents, ensuring that subtle morphological patterns are validated by institutional collective memory. This capability transforms the diagnostic workflow from an isolated event into a data-driven continuum, where the accuracy of a single case is reinforced by the synthesized evidence of thousands before it. In this way, institutional experience evolves from passive storage into an active, consultative knowledge system, an operational realization of the living library paradigm.

The framework is designed to meet the diverse needs of clinical practice, commercialization, research, and education by supporting a broad range of high-value use cases within a single, unified architecture. In the context of knowledge retrieval, PathoScribe has the potential to allow pathologists to explore diagnostic precedents and outcome patterns using conversational, natural language queries, and receiving synthesized, evidence-grounded responses drawn from institutional data, when deployed in the clinical workflow. For research and commercialization workflows, the system enables automated cohort construction by allowing users to define detailed inclusion and exclusion criteria in natural language and applying it across entire institutional archives, dramatically reducing the time and effort traditionally required for manual chart review. Within clinical decision support, PathoScribe provides data-driven immunohistochemistry (IHC) panel recommendations by integrating insights from similar historical cases, published literature, and practice guidelines, offering pathologists a transparent, context-aware starting point for diagnostic workups while preserving full human oversight [35].

Beyond these core applications, the framework also supports educational and communication-focused functions. It facilitates counterfactual "what-if" reasoning for trainees, converts narrative reports into standardized synoptic formats, and generates role-specific report translations tailored for oncologists, multidisciplinary teams, or patients. Additionally, it offers automated report summarization to streamline information transfer across clinical workflows [36, 37]. Each of these capabilities is implemented as a modular component within a cohesive system design, enabling flexible deployment and scalable integration into existing institutional infrastructures.

By demonstrating the capability of a single, LLM-driven framework to execute these diverse, high-stakes tasks across a large and heterogeneous dataset, PathoScribe represents a significant step towards realizing the potential of AI in digital pathology. Our work establishes a robust, scalable foundation for a future where pathology data is instantly accessible, interoperable, and directly translatable into clinical action. We anticipate that this unified approach will accelerate translational research, improve diagnostic consistency, and ultimately enhance patient care.

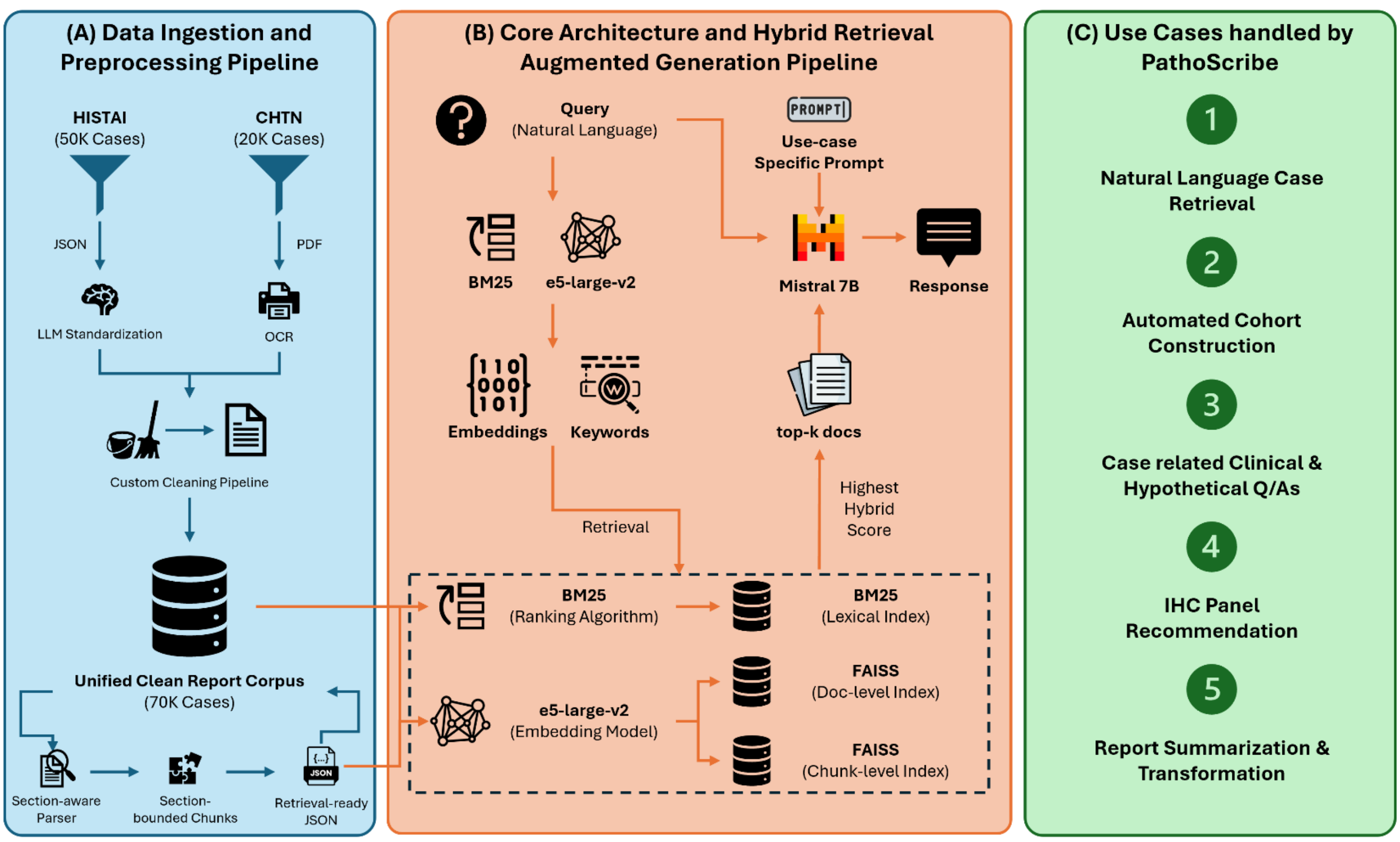

***Figure 1: Overview of the PathoScribe unified pathology intelligence framework. (A)*** *De-identified surgical pathology reports from multi-institutional archives undergo OCR, cleaning, and section-aware parsing to generate structured, retrieval-ready representations.* ***(B)*** *Hybrid retrieval combines dense document-level embeddings, dense chunk-level embeddings, and sparse BM25 lexical scoring into a unified relevance function for robust semantic case identification. Retrieved reports are assembled into a structured context and passed to an on-premises large language model (Mistral-7B-Instruct) within a retrieval-augmented generation (RAG) pipeline to ensure evidence-grounded responses.* ***(C)*** *The unified architecture supports diverse clinical workflows including natural language case retrieval, automated cohort construction, clinical question answering, IHC panel recommendation, and report transformation.*

## Results

### Dataset composition and corpus characteristics

To evaluate PathoScribe at scale, we assembled a large, heterogeneous corpus of 70,000 de-identified surgical pathology reports drawn from two independent, *multi-institutional archives*: HISTAI (n = 50,000) and the Cooperative Human Tissue Network (CHTN; n = 20,000) [38, 39]. CHTN archive provided reports spanning the years 2014 to 2025, primarily from the Midwest region, including institutions such as The Ohio State University (OSU), Case Western Reserve University (CWR), and the University of Pittsburgh (UP). The corpus included a broad multi-organ representation reflecting the diversity of routine surgical pathology practice, encompassing both oncologic and non-oncologic conditions. Although individual pathology report served as the unit of analysis in the present study, the framework preserves linkage between each report and its corresponding whole slide images (WSI) identifier, establishing a foundation for future multimodal integration. Collectively, this large and diverse corpus provided a realistic testbed for evaluating retrieval and language modeling performance across heterogeneous archival formats and reporting conventions, with applicability to both clinical workflows and research cohort development.

### PathoScribe outperforms keyword-based search for natural language case retrieval

Since the core functionalities of PathoScribe rely on accurately retrieving clinically relevant cases from free-text queries, we systematically evaluated its retrieval performance against conventional keyword-based search. We performed this analysis on a curated evaluation set of 100 pathology reports. Each report was paired with two query types: (i) a natural language context query describing the salient diagnostic and morphologic features of the case, and (ii) a keyword query consisting of discriminative terms expected to appear verbatim in the corresponding report.

For each target report (i.e., the ground-truth report from which the query was originally derived), we measured the rank position at which that report was retrieved from the full corpus. Retrieval performance was summarized using standard information retrieval metrics, Recall@1, Recall@5, and Recall@10, which quantify the proportion of queries for which the correct report appears within the top k returned results (Figure 2; Supplementary Table S1).

Context-aware semantic retrieval using PathoScribe demonstrated strong early-rank performance. Specifically, PathoScribe achieved Recall@1 of 0.8125, Recall@3 of 0.90625, and Recall@10 of 1.0, indicating that the correct report was almost always ranked within the top few results and always within the top 10. This performance is consistent with clinically realistic browsing depth. In contrast, traditional keyword-based retrieval performed substantially worse (Recall@1 0.1875, Recall@3 0.3125, Recall@10 0.40625), even when provided with idealized, handcrafted term lists, highlighting the limitations of exact-match search in the presence of linguistic variability in pathology reporting.

**Natural Language Case Retrieval**

*“Find a case of hepatocellular carcinoma treated by partial hepatectomy with background cirrhosis, micro-vascular invasion, and elevated AFP.”*

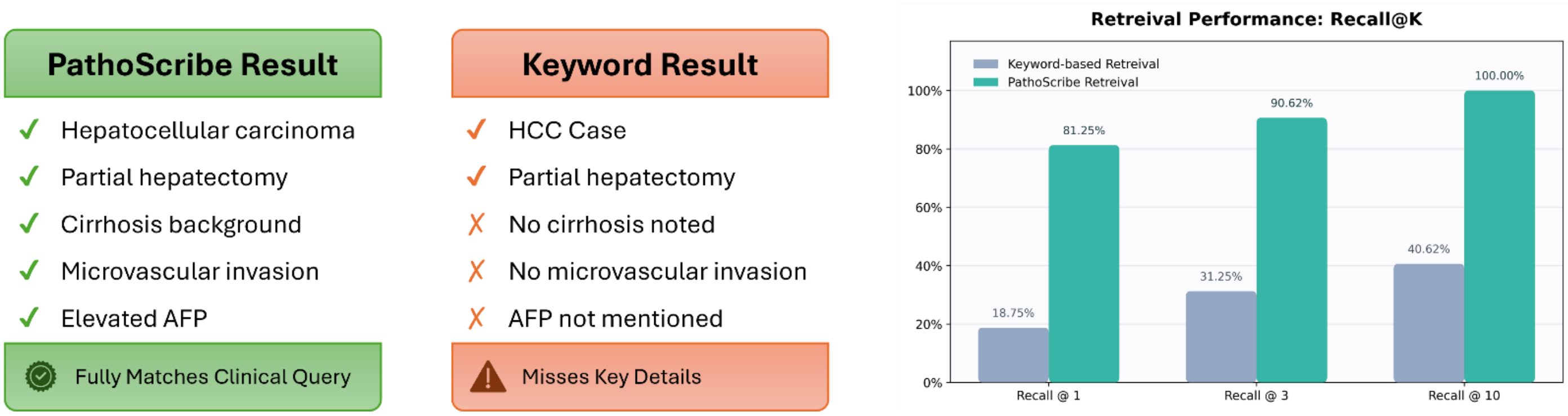

***Figure 2: Natural language case retrieval with PathoScribe.*** *Comparison of semantic retrieval using PathoScribe versus conventional keyword-based search for clinical case matching.* ***Left,*** *PathoScribe interprets the full clinical context of the query and retrieves highly concordant cases, achieving comprehensive alignment with key clinical attributes.* ***Middle,*** *keyword-based retrieval relies on surface term matching and frequently misses critical contextual details.* ***Right,*** *quantitative evaluation demonstrates consistently higher Recall@K for PathoScribe across retrieval depths, indicating superior early-hit performance and overall case relevance. Together, these results highlight the advantage of context-aware retrieval for pathology case discovery and downstream clinical decision support.*

To identify the optimal embedding backbone for dense retrieval, we systematically compared multiple candidate models and pooling strategies on the curated retrieval benchmark described above. The selected backbone was subsequently used across all downstream use cases within the PathoScribe framework to ensure consistency of representation.

Among the evaluated models, e5-large-v2 demonstrated superior performance in ranking the target report highest in response to natural language queries, achieving the strongest early-rank recall and overall retrieval accuracy. Detailed comparative results across all tested backbones are provided in Supplementary Tables 2–4.

**PathoScribe enables rapid, automated cohort construction from free-text criteria**

Cohort construction in pathology traditionally relies on labor-intensive manual chart review, often requiring hundreds of hours of expert effort. This process can delay cohort assembly for months, increase study costs, and limit the feasibility of exploratory or hypothesis-generating research. As a result, many potentially informative clinical and translational studies remain underpowered or are never pursued due to the practical burden of assembling well-defined patient cohorts in a timely manner [40-43].

PathoScribe addresses this critical bottleneck by automating cohort identification directly from free-text inclusion and exclusion criteria. By leveraging semantic retrieval and language understanding, the framework can autonomously identify and construct research-ready cohorts within minutes, enabling rapid feasibility assessment and lowering the barrier to testing less-likely or exploratory hypotheses that would otherwise be prohibitively time-consuming to evaluate.

To validate automated cohort construction, we defined 10 distinct cohort specifications using clinician-authored free-text inclusion and exclusion criteria and audited model decisions through blinded human review. End-to-end cohort assembly was efficient across all specifications, requiring a mean of 9.2 minutes per cohort (median: 5.9 minutes; range: 1.0–36.8 minutes) on a cost-effective, commonly available NVIDIA A100 40GB GPU (~$2–$3 per hour on-demand in major cloud platforms and one-time cost of ~$8k–$12k for on-premise hardware), representing orders-of-magnitude lower cost than manual chart review and demonstrating the practical scalability of the framework for real-world use (Supplementary Table S8, Supplementary Methods S1.6). From the resulting cohorts, we randomly sampled 60 cases, including both included (n = 30) and excluded (n = 30) by PathoScribe, reflecting a realistic clinical validation scenario in which both false inclusions and missed eligible cases are critical.

Two independent medical-student reviewers evaluated all sampled cases using a binary correctness rubric (1 = correct inclusion/exclusion decision; 0 = incorrect) and provided free-text justifications to document reasoning and identify discrepancies. Across this evaluation, PathoScribe achieved 91.34% overall accuracy for eligibility classification (Figure 3). Notably, no eligible cases were incorrectly excluded, indicating strong sensitivity in preserving clinically relevant cases during cohort assembly.

**Automated Cohort Construction**

**Inclusion Criteria:**
- Greater than 18 years old
- Histologically confirmed Stage I-III Invasive Lung Adenocarcinoma
- Underwent surgical resection between 2008 to 2015

**Exclusion Criteria:**
- With previous lung cancer
- With other cancers or life-threatening conditions
- Mucinous adenocarcinoma

**PathoScribe Result**

| Patient ID | Decision |
|---|---|
| P-001 | Exclude |
| P-002 | Exclude |
| P-003 | Include |
| P-004 | Include |
| P-005 | Exclude |

**Evaluates each case in the database and creates a cohort.**

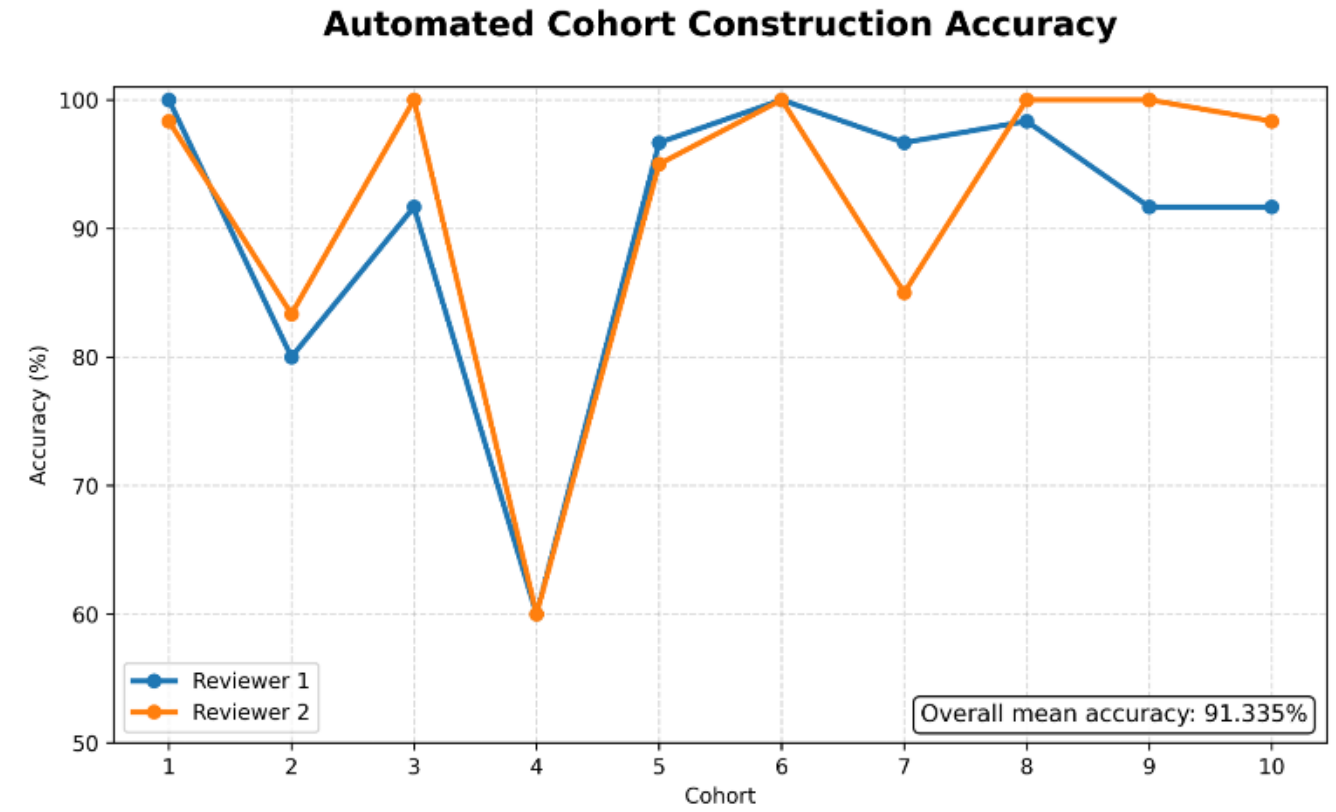

***Figure 3: Automated cohort construction with PathoScribe.*** *Overview of PathoScribe for automated patient cohort identification from natural language eligibility criteria.* ***Top,*** *the system semantically evaluates each case against study criteria, in contrast to conventional approaches that frequently fail to capture full contextual requirements.* ***Bottom left****, example output showing patient-level inclusion and exclusion decisions generated by PathoScribe to assemble the final cohort.* ***Bottom right,*** *quantitative evaluation demonstrates high agreement with expert reviewers across cohorts, with a mean accuracy exceeding 91%. These results highlight the reliability of PathoScribe for scalable, criteria-driven cohort construction in clinical research workflows.*

### PathoScribe delivers accurate answers to case-specific and hypothetical clinical questions

Answering targeted clinical questions from pathology reports can be time-consuming, often requiring manual review of lengthy narratives to extract relevant findings. The challenge becomes even greater when clinicians or researchers seek to test specific hypotheses, identify disease precedents across patient cohorts, or explore counterfactual "what-if" scenarios. These tasks typically demand substantial expert effort and iterative database searches. PathoScribe automates this process by generating retrieval-grounded responses directly from pathology reports and by synthesizing evidence-informed answers to hypothetical queries using both institutional cases and curated external knowledge sources.

To assess answer quality for case-specific clinical questions, we evaluated retrieval-grounded responses using a five-point Likert scale (5 = best; 1 = worst) scored independently by two human reviewers. A total of 64

question–answer pairs were assessed. Scores were strongly concentrated at the upper end of the scale, with 71.9% receiving a score of 5 and 90.6% receiving scores of 4–5. The mean score was 4.56, with a 95% confidence interval of [4.35, 4.78], indicating consistently high perceived answer quality (Figure 4).

**Case Related Question Answers**

*"I'm looking for a case with a patient who is male and between 65-80 years old and has a high grade urothelial carcinoma of the renal pelvis. Explain the procedure that the patient underwent."*

**PathoScribe Result**

**Best Matching Case Number:** ABCXXXX

**Procedure details:**

- Main operative procedure was nephroureterectomy.

**Additional treatment:**

- No specific systemic therapy.

**Margin status:**

- Margins were uninvolved.

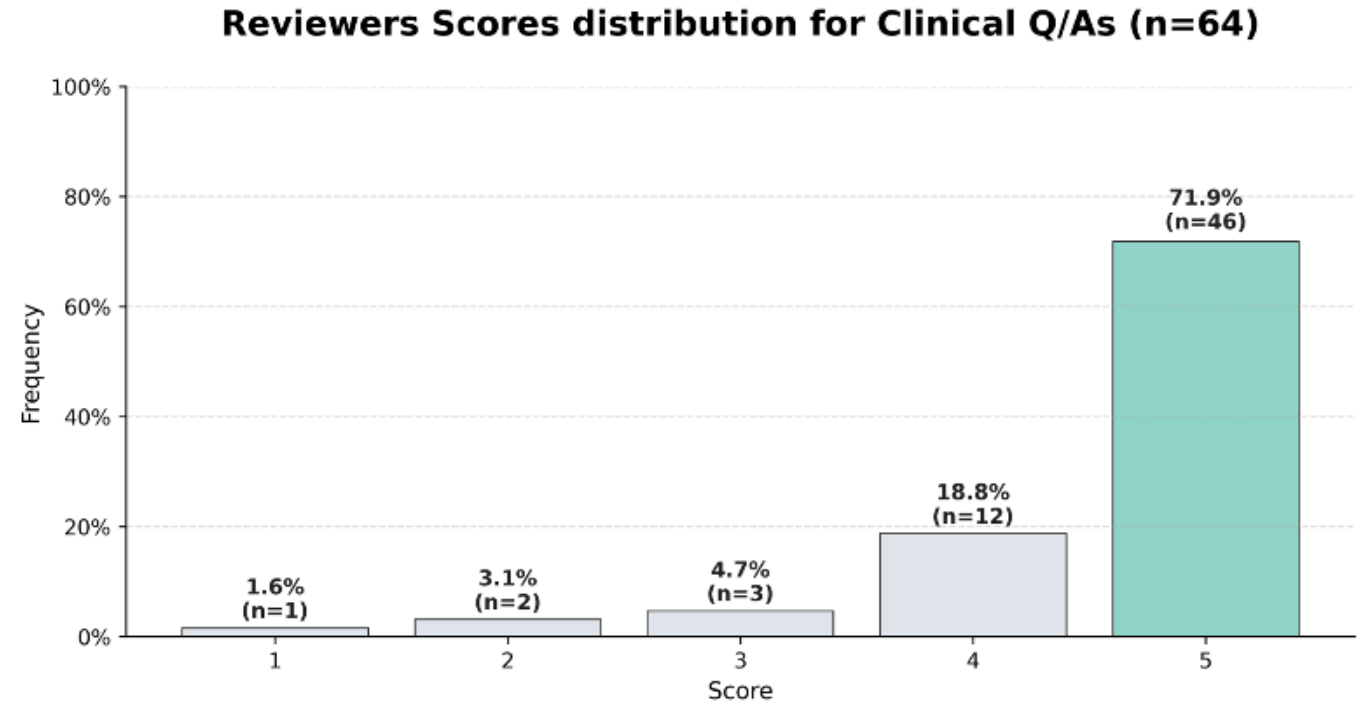

***Figure 4: Reviewer score distribution for case-specific clinical question answering.*** *Distribution of independent reviewer scores (1–5 Likert scale; 5 = best) for PathoScribe responses to case-specific clinical questions derived from pathology reports (n = 64). Most responses were rated highly, with 71.9% receiving a score of 5 and 90.6% receiving scores of 4–5 overall. The mean score was 4.56 (95% CI: 4.35–4.78). These results demonstrate consistent answer quality and strong perceived clinical relevance for retrieval-grounded case-level question answering.*

We further evaluated PathoScribe on hypothetical "what-if" reasoning tasks through blinded human review of 100 generated responses. The evaluation was organized into eight review packets (25 cases each, sampled with replacement such that every case appeared in at least one packet) and assessed by four independent reviewers, each scoring two packets. Reviewers assigned a five-point correctness score (5 = correct; 1 = incorrect) and provided qualitative feedback. Across all reviewed answers, PathoScribe achieved a mean correctness score of 4.85/5, with the distribution heavily skewed toward fully correct responses (Figure 5).

**Hypothetical Question Answers**

*"What if the carcinoma had only invaded the inner half of the myometrium instead of the outer one-third (70% depth)?"*

**PathoScribe Result**

"If the carcinoma had only invaded the inner half of the myometrium, it would be considered a low-stage endometrial carcinoma."

--------------------------------------------------------------------------------

***Clinical Guidance:***

- *Less myometrial invasion (<50% or inner half) corresponds to a lower FIGO stage.*

***Assumptions:***

- *Assumes endometroid type (most common)*
- *Stage determined solely by invasion depth, lacking other prognostic factors (e.g., lymph node involvement)*

***Sources:***

- *Path report case number ABCXXXX chunk number 4623*

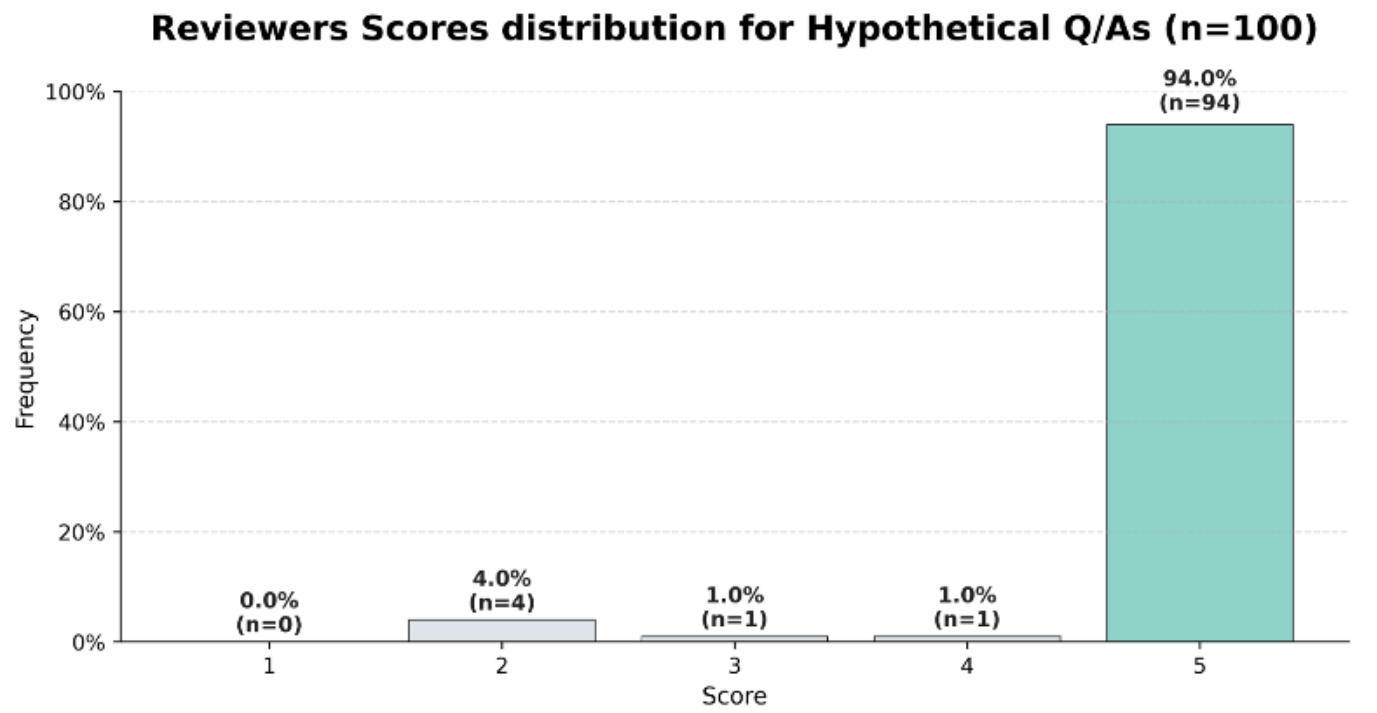

***Figure 5: Reviewer score distribution for hypothetical ("what-if") question answering.*** *Distribution of blinded reviewer scores (1–5 scale; 5 = fully correct) for PathoScribe responses to hypothetical counterfactual clinical questions (n = 100). Bars indicate percentage frequency for each score category. The distribution is strongly skewed toward the highest rating, with 94% of responses receiving a score of 5 and a mean correctness score of 4.85/5. These findings indicate high perceived correctness and reasoning quality in retrieval-grounded, counterfactual educational scenarios.*

**PathoScribe improves information-efficient IHC panel selection**

Selecting an initial IHC panel is fundamentally an information-gain decision under diagnostic uncertainty: the goal is to identify a compact set of markers that most efficiently distinguishes among the leading differential diagnoses based on the histomorphologic findings and clinical context. The panel typically includes both positive and negative markers, confirming the primary diagnostic consideration while excluding potential morphologic mimickers. In routine practice, this process depends heavily on individual experience and comparison with prior cases if applicable, which can introduce variability and inefficiency. PathoScribe operationalizes this task through a retrieval-augmented, candidate-constrained recommendation strategy that grounds panel selection in analogous institutional cases and patient-specific context.

To ensure a fair and leakage-free evaluation, all IHC-related content was removed from each report prior to model inference, preventing exposure to prior staining decisions. We compared PathoScribe's recommendations against a direct LLM generation baseline (Mistral 7b) across 270 cases spanning eight organ systems. Performance was assessed using paired bootstrap estimates (2,000 resamples), with results summarized in Figure 6 and Supplementary Table S5 for both overall and organ-stratified macro metrics.

# IHC Panel Recommendation

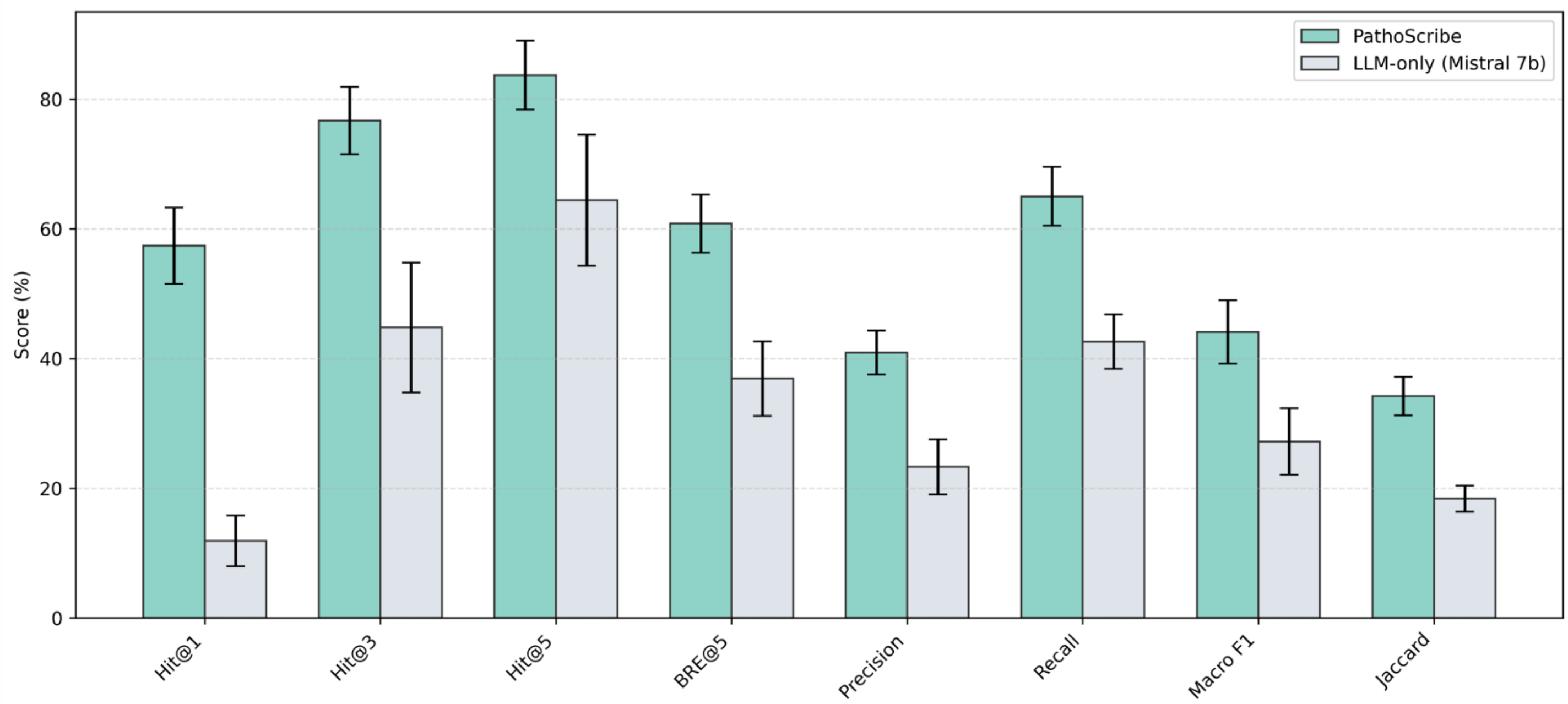

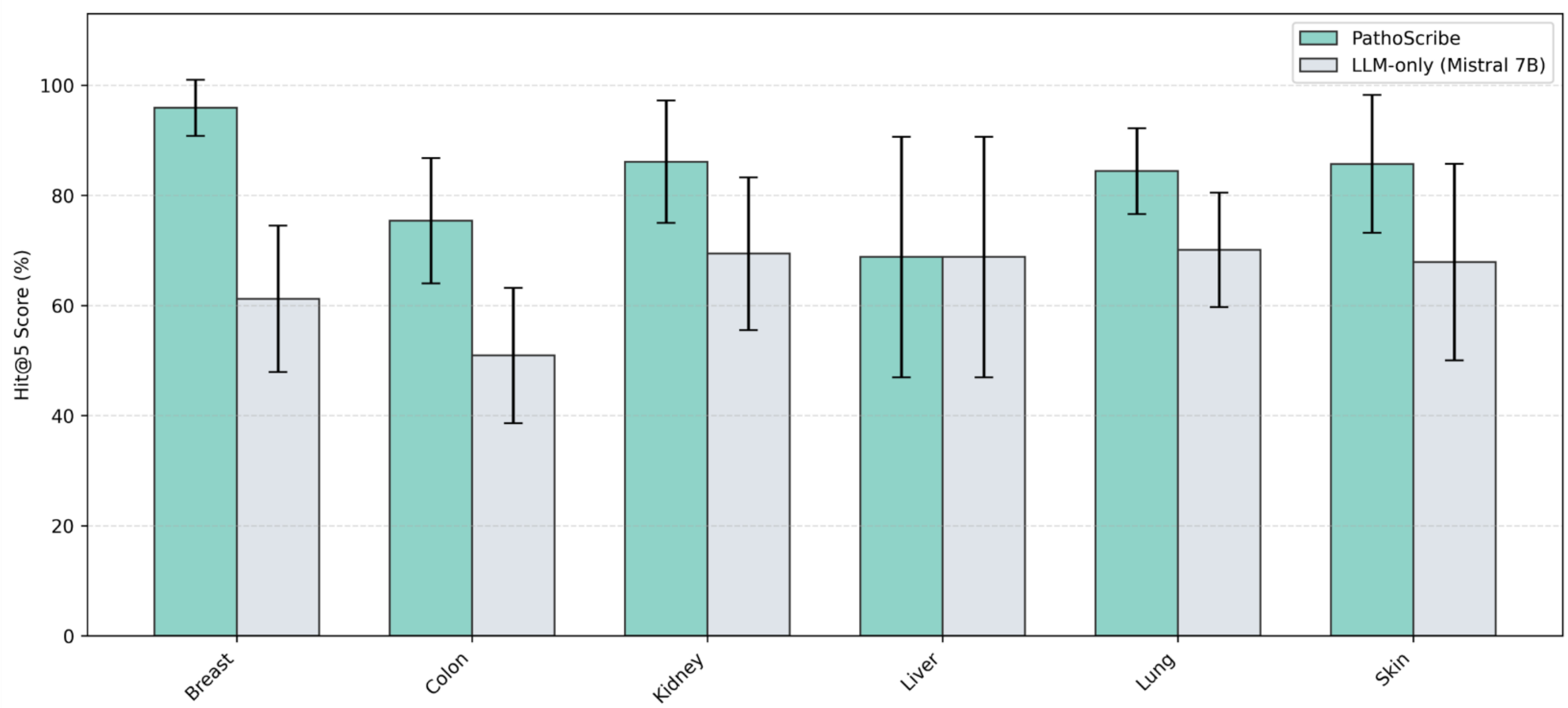

***Figure 6: IHC panel recommendation performance compared with direct LLM generation. Top,*** *overall IHC panel recommendation performance across ranking and set-agreement metrics, comparing PathoScribe (retrieval-augmented) with LLM-only generation (Mistral-7B). Metrics include Hit@1, Hit@3, Hit@5, budgeted recall at 5 (BR@5), precision, recall, macro F1, and Jaccard similarity.* ***Bottom,*** *organ-specific Hit@5 performance across breast, colon, kidney, liver, lung, and skin cases. Error bars denote 95% bootstrap confidence intervals (2,000 resamples). PathoScribe demonstrates substantial improvements in early-hit*

*performance and marker-set agreement, indicating that retrieval-grounded recommendations more closely align with clinically ordered IHC panels than unguided LLM generation.*

PathoScribe demonstrated substantial improvements in early-hit performance and case-level utility. Hit@k measures early-hit performance and is defined as the proportion of cases in which at least one ground-truth IHC marker appears within the top k recommended markers. Budgeted recall at 5 (BR@5) quantifies coverage under practical staining constraints and is defined as the recall achieved when only the top five recommended markers are considered. Precision is the fraction of recommended markers that are present in the reference panel, recall is the fraction of reference markers successfully retrieved by the recommendation, and F1 score is the harmonic mean of precision and recall. Jaccard similarity measures set-level overlap between the recommended and reference marker panels and is computed as the size of the intersection divided by the size of the union of the two sets.

Upon evaluation using these metrics, Hit@1 increased from 11.9% to 57.4% (4.8×), with consistent gains at Hit@3 (76.7) and Hit@5 (83.7), alongside markedly higher budgeted recall within a five-marker budget (BR@5: 60.8). These ranking improvements translated into stronger marker-set agreement, reflected in higher precision (40.9), recall (65.0), F1 score (46.1), and Jaccard similarity (34.2), across major organ systems, indicating more clinically aligned and information-efficient IHC panel recommendations.

**Prompt-controlled report transformation enables high-fidelity, audience-specific outputs**

Beyond retrieval and decision support, PathoScribe implements prompt-controlled transformation modules that restructure narrative pathology reports into task- and audience-specific formats. By varying only the prompt, the same source report can be rendered as: (i) a structured CAP-style synoptic report with extracted key elements and flagged missing fields, (ii) a concise clinician-facing summary highlighting key findings and actionable results, or (iii) role-adapted versions tailored for oncologists, tumor boards, or patient-friendly communication. This design enables flexible downstream use without requiring task-specific model retraining.

We evaluated the transformation module on n = 50 pathology reports across five downstream renderings: (i) CAP-style synoptic conversion, (ii) concise clinician-facing summarization, (iii) oncologist-facing reports, (iv) tumor-board narratives, and (v) patient-friendly summaries. Because routine clinical archives rarely contain paired human rewrites for all formats, evaluation focused on complementary fidelity and readability metrics: (a) lexical overlap with available reference rewrites (ROUGE-1/2/L and BLEU-4), (b) semantic alignment (embedding cosine similarity and PathBERTScore), (c) source-grounded content retention (CSR fact coverage recall), (d) faithfulness guardrails (numeric precision and auditor-supported rate), and (e) readability (Flesch–Kincaid grade level and Flesch Reading Ease) (Figure 7, Supplementary Tables S6-7).

## Report Transformation

**Report Summarization & Translation Performance**

***Figure 7: Report transformation performance across summarization and translation tasks.*** *Performance of PathoScribe across five report transformation outputs: concise clinical summary, oncologist-facing report, tumor board–facing report, CAP-style synoptic conversion, and patient-friendly report. Metrics include lexical overlap (ROUGE-1/2/L F1, BLEU-4), semantic similarity (embedding cosine similarity, PathBERTScore), and readability-oriented adaptation. Error bars represent 95% confidence intervals across n = 50 reports. Clinician- and tumor board–facing outputs demonstrate strong semantic alignment and content retention, while CAP-style synoptic reports achieve near-perfect semantic fidelity. Patient-friendly reports maintain high semantic similarity while optimizing accessibility. These results demonstrate that prompt-controlled transformation preserves diagnostic fidelity while enabling audience-specific restructuring.*

Across clinician- and tumor-board-facing outputs, PathoScribe demonstrated strong lexical and semantic alignment with reference rewrites (ROUGE-1: 0.61–0.65; PathBERTScore: 0.72–0.78) while retaining a substantial proportion of extractable clinical facts (CSR recall: 0.60–0.70). The oncologist-facing configuration achieved the highest content retention (CSR recall: 0.70), accompanied by high auditor support (0.92) and perfect numeric precision in this cohort. Patient-facing summaries met accessibility targets (Flesch–Kincaid grade 5.86; Reading Ease 69.4) while maintaining strong faithfulness (numeric precision 1.00; auditor-supported 0.82). CAP-style synoptic outputs remained highly semantically aligned with available references (PathBERTScore: 0.98) and preserved source facts (CSR recall: 0.66), consistent with high-fidelity structured re-expression.

## Methods

### Data Preprocessing

The HISTAI and CHTN reports were stored in heterogeneous formats and exhibited substantial variability in structure, language, and content. To harmonize these sources, we implemented cohort-specific cleaning and preprocessing pipelines followed by standardized normalization procedures, resulting in a unified corpus of approximately 70,000 reports suitable for downstream analysis.

Reports from the HISTAI archive were initially stored in structured, JSON-like format. These files were processed by passing each object sequentially to an LLM to transform the structured data into a continuous, narrative text representation, ensuring the preservation of the full report content.

Reports from the CHTN archive were provided as Portable Document Format (PDF) files and were converted to text using an optical character recognition (OCR) engine [44]. The OCR workflow renders each PDF page to an image, applies OCR to each page, and merges page text into a single document text file while retaining page boundaries for traceability. A post-OCR cleaning step normalizes whitespace, repairs hyphenation artifacts across line breaks, removes common pagination markers and repetitive boilerplate, and produces a clean, contiguous text representation suitable for downstream sectioning and retrieval indexing.

A section-aware parser then identifies common report components using regular expressions over standard headings (e.g., diagnosis, microscopic description, gross description, comment, and immunohistochemistry). Cleaned documents are converted into retrieval-ready line-delimited JSON containing document-level records, section-bounded text chunks, section labels, and short summaries. Detailed parameter settings for OCR, cleaning, and chunking are provided in the Supplementary Methods.

**Overview of PathoScribe Framework**

The PathoScribe framework was deployed on on-premises GPU infrastructure to ensure data security and regulatory compliance. LLM inference for downstream generation tasks was performed using the Mistral-7BInstruct model hosted on a local inference server, with model weights stored on institutional storage and all requests confined to the institutional network [45]. The system was designed with a modular API architecture to facilitate integration with existing laboratory information systems (LIS) and electronic health record (EHR) platforms through RESTful endpoints / standard web interfaces, while safeguarding all protected health information (PHI) and model execution within secure, controlled computing environments.

**Embedding Model and Semantic Encoding**

The default embedding backbone for semantic retrieval was e5-large-v2 [46]. Both document-level representations and section-bounded chunks derived from the preprocessing pipeline were embedded. Input text is tokenized with padding and truncation to a maximum length of 512 subword tokens and passed through the encoder to obtain final hidden states of shape $(B, L, d)$, where $B$ is batch size, $L$ is sequence length, and $d$ is the dimension of embeddings. Fixed-length embeddings are computed using masked mean pooling over non-padding tokens followed by row-wise $l_2$ normalization. Document-level embeddings and chunk-level embeddings are stored in line-delimited JSON index tables that maintain traceability between dense vectors and source text.

**Vector Database and Indexing**

Encoded embeddings were indexed using FAISS (Facebook AI Similarity Search) for efficient similarity search in high-dimensional spaces [47, 48]. Because all embeddings are $l_2$-normalized, we use inner product as the similarity measure, which is equivalent to cosine similarity under unit-norm constraints. We construct two exact dense indices using an inner-product flat index: one over the doc-level embeddings and one over the chunk-level embeddings. In parallel, we build a sparse lexical index using BM25 [49]. Document text is tokenized using whitespace splitting and indexed with a standard BM25 implementation, and the list of report identifiers is saved alongside the BM25 model. This yields three complementary retrieval backends: a doc-level dense FAISS index, a chunk-level dense FAISS index, and a BM25 index.

**Hybrid Scoring Formulation**

The hybrid retrieval module combines dense doc-level signals, dense chunk-level signals, and sparse BM25 scores into a single relevance score for each report [50]. Let $q \in \mathbb{R}^d$ denote the normalized query embedding retrieved by passing the query text to the embedding model. Let $D(r)$ be the set of embedding indices corresponding to doc-level representations for report $r$, and let $C(r)$ be the set of embedding indices for chunks belonging to $r$.

We define the doc-level similarity as the maximum cosine similarity between the query and any doc-level embedding for $r$:

$$s_{doc}(r) = \max_{\mathrm{j \in D(r)}} \langle q, d_j \rangle,$$

where $d_j$ is the normalized embedding of document $j$ and $\langle \cdot, \cdot \rangle$ denotes the inner product.

Similarly, the chunk-level similarity is

$$s_{chunk}(r) = \max_{k \in C(r)} \langle q, c_k \rangle,$$

with $c_k$ denoting the normalized embedding of chunk $k$. The chunk $k^*$ attaining this maximum is treated as the best chunk for r for display and error analysis.

For BM25, let $s_{bm25}^{raw}(r)$ denote the unnormalized BM25 score of report $r$ for the tokenized query. We normalize BM25 scores to the range $[0, 1]$ on a per-query basis:

$$s_{bm25}(r) = \begin{cases} \dfrac{s_{bm25}^{raw}(r)}{max_{r'}\, s_{bm25}^{raw}(r')}, & max_{r'}\, s_{bm25}^{raw}(r') > 0 \\ 0, & otherwise \end{cases}$$

The final hybrid relevance score for a report $r$ is a linear combination of these components:

$$score(r) \; = \; \alpha_{doc} s_{doc}(r) \; + \; \alpha_{chunk} s_{chunk}(r) \; + \; \alpha_{bm25} s_{bm25}(r),$$

with $\alpha_{doc} + \alpha_{chunk} + \alpha_{bm25} = 1$. Default weights are reported in Supplementary Materials. Reports are ranked in descending order of $score(r)$, and the top-k are passed to downstream LLM-based workflows.

### Large Language Model Configuration

Mistral-7B-Instruct served as the generation backbone for all NLU and generation tasks within PathoScribe. The model was hosted on an on-premises GPU server using a high-throughput inference engine, with weights loaded from local storage and all traffic restricted to the institutional network. We configured the model with an 8,192-token context window and used standard sampling-based decoding for generative tasks to balance response diversity with consistency. All prompts were structured to include explicit system instructions that guided the model to ground responses in retrieved evidence and avoid unsupported speculation.

### Retrieval-Augmented Generation Pipeline

PathoScribe implements a hybrid retrieval approach that combines dense semantic search with sparse lexical matching, i.e., integrating meaning-based search with traditional keyword matching, to maximize both semantic relevance and lexical precision for improved relevance and precision of information retrieval. When processing a natural language query, the system encodes the query using e5-large-v2 with the same tokenization and pooling strategy used for corpus embedding, producing a single normalized query vector $q$.

This query vector is then used to perform dense search against both FAISS indices. For the doc-level index, we retrieve the top-k document candidates by cosine similarity. For the chunk-level index, we retrieve the top-k chunk candidates and record, for each report, the highest-scoring chunk and its section label as a localized relevance signal.

In parallel, we apply BM25 to the whitespace-tokenized corpus, scoring indexed reports and selecting a top-k subset by lexical relevance; BM25 scores are normalized to the range $[0, 1]$ as described above.

Dense and lexical signals are then aggregated per report identifier using hybrid score. For each report, we track (i) the maximum dense similarity among its doc-level matches, (ii) the maximum dense similarity among its chunk-level matches along with the identity and section type of the best-scoring chunk, and (iii) the normalized BM25 score if the report appears in the BM25 top-k set. Reports are ranked in descending order of this combined score, and the top results are used as the retrieval set for downstream LLM-based workflows. For interactive search and debugging, we additionally surface a short snippet from each report together with the text of its best-scoring chunk section.

**Context Assembly and Prompt Construction**

For use cases that require generative responses, the top-ranked reports from the hybrid retrieval module are used to construct a structured context prompt for Mistral-7B-Instruct. The controller assembles a prompt that begins with a fixed system instruction describing the model's role and grounding requirements, followed by a block containing the retrieved report texts and a final block containing the user's query and task-specific response instructions.

To balance context coverage with the model's fixed context window, we include the highest-ranked subset of retrieved reports and truncate individual report texts as needed so that the combined context, plus instructions and query, remains within the 8,192-token limit. In practice, this typically involves including the most informative parts of each report and eliding long, repetitive segments when necessary. A generic template for case retrieval tasks is as follows:

```
[SYSTEM INSTRUCTION]
You are an expert pathology assistant. Your role is to synthesize information exclusively from the provided
pathology reports. Base all responses on evidence found in these reports and cite specific cases when
making claims. Do not introduce information beyond what is present in the retrieved documents.
[RETRIEVED REPORTS]
Report 1: [Truncated text of report 1]
Report 2: [Truncated text of report 2]
. . .
Report N : [Truncated text of report N ]
[USER QUERY]
{user's natural language query}
[RESPONSE INSTRUCTION]
Provide a comprehensive answer based on the reports above, citing specific case identifiers when
referencing information.
```

This prompt engineering approach ensures that the LLM's responses remain grounded in institutional data while leveraging its language understanding capabilities to synthesize coherent, clinically relevant answers.

**Use Case-Specific Implementations**

**Use Case 1: Natural Language Case Retrieval**

The case retrieval module enables pathologists to explore institutional precedents and diagnostic patterns through conversational queries such as "What is the common outcome of patients with stage III colon adenocarcinoma?" The system processes such queries through the hybrid retrieval pipeline integrating meaning-based search with traditional keyword matching, returning both, a synthesized natural language response generated by Mistral-7B based on the retrieved reports, and the list of top-k relevant case reports with their identifiers and key metadata. This dual output format allows users to read the AI-generated summary for rapid insight while maintaining access to source reports for verification and detailed review. Practically, the retrieved example reports may assist pathologists in drafting their own reports, while AI-generated insights provide context and educational support for refining the final wording.

**Use Case 2: Automated Research Cohort Construction**

Cohort construction represents one of the most time-intensive tasks in clinical research, traditionally requiring manual chart review that can take weeks or months for large institutional databases. PathoScribe automates this process by accepting free-text inclusion and exclusion criteria expressed in natural language, such as:

"Include: Adult patients (>18 years) with biopsy-proven invasive ductal carcinoma of the breast, ER-positive, HER2-negative, with lymph node involvement. Exclude: Patients with metastatic disease at diagnosis, prior history of breast cancer, or neoadjuvant therapy."

The system converts these criteria into a structured evaluation prompt and processes each report in the database independently through Mistral-7B. For each case, the model receives a prompt containing:

```
[SYSTEM INSTRUCTION]
You are tasked with determining whether a pathology case meets specific inclusion and exclusion criteria
for a research cohort. Evaluate the case based solely on information present in the report. Return your
decision in JSON format.
[COHORT CRITERIA]
{user-provided inclusion and exclusion criteria}
[PATHOLOGY REPORT]
{text of individual report}
[OUTPUT FORMAT]
{"case_number": "XXXXX", "decision": 0 or 1, "rationale": "brief explanation"}
Where decision=1 means INCLUDE and decision=0 means EXCLUDE.
```

The LLM evaluates each case individually, returning a JSON object containing the case identifier, binary inclusion decision (0/1), and brief rationale. This approach scales efficiently across tens of thousands of reports and handles complex criteria involving logical operators (AND/OR/NOT) through the model's natural language understanding.

**Use Case 3: Educational "What-If" Scenarios**

PathoScribe includes an educational module that enables trainees to explore counterfactual reasoning by uploading a patient report and posing hypothetical modifications such as "What if this patient had been 20 years younger?" or "What if the tumor showed signet ring cell morphology instead of glandular architecture?" The system leverages both its pretrained medical knowledge and retrieval of similar historical cases to generate pedagogically valuable responses that discuss how diagnostic workup, differential diagnosis, or prognosis might differ under the altered conditions. This use case is explicitly designed as an educational and exploratory tool, not for clinical decision-making, and responses include appropriate disclaimers regarding the hypothetical nature of the analysis.

**Use Case 4: IHC Panel Recommendation**

Immunohistochemistry (IHC) plays a critical role in diagnostic pathology, yet selecting optimal antibody panels requires integration of morphologic features, clinical context, knowledge of diagnostic algorithms, and reflex biomarkers for treatment indications. PathoScribe assists pathologists by recommending IHC panels based on patient demographics, tumor characteristics, and clinical decision goals (e.g., subtyping, prognostic markers, treatment selection).

When a user inputs case details, the system retrieves semantically similar historical cases from the institutional database and examines the IHC panels previously ordered in analogous diagnostic scenarios. The LLM then synthesizes this institutional experience with its pretrained knowledge of IHC utility and diagnostic algorithms to generate a ranked list of suggested antibodies, each accompanied by a brief rationale explaining its diagnostic relevance. The recommendations are explicitly labeled as advisory and subject to pathologist verification before clinical use. This approach provides a transparent, evidence-grounded starting point that accelerates diagnostic workup while preserving full human oversight.

**Use Case 5: Report Standardization and Communication**

Beyond retrieval and decision support, PathoScribe implements several natural language transformation modules to enhance report utility across different clinical contexts:

**Synoptic Report Conversion:** Narrative pathology reports are converted into structured synoptic format following College of American Pathologists (CAP) protocol templates. The LLM extracts key diagnostic elements (tumor size, margins, lymph node status, staging parameters) and populates standardized synoptic fields, flagging any missing elements that require manual completion.

**Report Summarization:** Lengthy pathology reports, particularly those involving complex resection specimens, are condensed into concise summaries highlighting key diagnostic findings, staging information, and clinically actionable results. This facilitates rapid information transfer during multidisciplinary tumor boards and clinical handoffs.

**Role-Specific Translation:** PathoScribe generates tailored report versions optimized for different audiences, including oncologist-focused summaries emphasize staging, prognostic markers, and treatment-relevant biomarkers; tumor board presentations provide structured case summaries suitable for multidisciplinary discussion; patient-friendly reports translate medical terminology into accessible language at an appropriate health literacy level (6th-8th grade reading level) while preserving diagnostic accuracy.

Each transformation module employs specialized prompts that instruct Mistral-7B to restructure content according to the target format and audience while maintaining fidelity to the original diagnostic information.

**Discussion**

Pathology practice remains fundamentally text-driven, yet most computational pathology advances to date have focused primarily on WSI analysis. Here we introduce PathoScribe, a unified pathology intelligence framework that brings semantically grounded natural language understanding to large-scale pathology archives. Across multiple clinically relevant workflows, including case retrieval, cohort construction, question answering, IHC recommendation, and report transformation, the framework demonstrates that a single retrieval-augmented architecture can support diverse, high-value tasks within routine pathology operations.

A central contribution of PathoScribe is its ability to enable reliable natural language retrieval from unstructured pathology reports. Traditional keyword-based systems are inherently brittle in the face of synonymy, variable phrasing, and narrative heterogeneity that characterize real-world reports. By contrast, the hybrid dense–sparse retrieval strategy consistently identified clinically relevant cases from free-text queries, supporting more natural and efficient archive exploration. Importantly, the system was able to place the most relevant results near the top of the list, which is essential for practical use in real-world settings. These findings highlight a fundamental limitation of traditional lexical matching in pathology archives: clinically relevant reports often fail to share exact phrasing with user queries, despite clear semantic alignment. By contrast, semantic retrieval within PathoScribe robustly captures contextual meaning, enabling reliable identification of relevant cases from natural language descriptions despite variations in terminology or individual wording preferences. While we expect the current architecture to remain robust as institutional archives scale to millions of reports, we recognize a known limitation of cosine-similarity–based retrieval in large knowledge bases: high-scoring chunks can become concentrated within a small subset of documents, potentially reducing result diversity. Future iterations will incorporate maximal marginal relevance (MMR)–based diversification to mitigate this effect and improve coverage across heterogeneous cases.

Automated cohort construction represents another high-impact application addressed by PathoScribe. Manual cohort assembly is widely recognized as a major bottleneck in clinical research, often requiring hundreds of hours of expert chart review. Our results demonstrate that free-text eligibility criteria can be operationalized at scale with high agreement to human reviewers. Nevertheless, reviewer feedback highlighted two primary sources of residual error: (i) hallucinated eligibility attributes when required evidence was absent from the pathology report, and (ii) ambiguity-driven disagreements when inclusion or exclusion criteria were underspecified or open to interpretation. Both issues are tractable. First, performance is expected to improve substantially when the system is integrated with full patient context, including clinical notes, encounter summaries, and longitudinal chart data, rather than relying solely on pathology reports. Second, although the system accepts free-text criteria, clearer and more structured specifications of eligibility rules can further reduce ambiguity. Together, these enhancements should strengthen reliability for high-stakes research cohort assembly.

For IHC panel recommendation, PathoScribe frames marker selection as an information-gain problem grounded in institutional precedent. Importantly, the system is designed as a decision-support aid rather than a prescriptive tool, providing transparent rationales and preserving full pathologist oversight. Future prospective reader studies will be important to measure real-world impact on diagnostic efficiency and cost-effectiveness.

The prompt-controlled transformation module further demonstrates the flexibility of the unified architecture. By varying only the prompt, PathoScribe can generate synoptic reports, clinician summaries, tumor board narratives, and patient-friendly explanations while maintaining strong content fidelity. This capability has practical implications for reducing documentation burden and improving cross-disciplinary communication. However, broader validation across additional organ systems, institutions, and reporting styles will be necessary to fully characterize generalizability.

This study has several limitations. First, although the corpus is large and multi-institutional, it is still predominantly drawn from Midwestern U.S. practice settings; external validation in more geographically and stylistically diverse datasets will be important. Second, the current system operates primarily on text reports and does not yet exploit the rich visual information available in paired whole-slide images. Third, as with all LLM-based systems, outputs remain sensitive to prompt design and retrieval quality, underscoring the importance of continued guardrail development and human-in-the-loop oversight. Finally, prospective clinical deployment studies are needed to quantify real-world workflow impact, user trust, and downstream patient outcomes.

Despite these limitations, PathoScribe establishes a scalable foundation for natural language–driven pathology intelligence. By unifying retrieval, reasoning, and report transformation within a single modular framework, the system moves beyond task-specific prototypes toward a more general-purpose pathology assistant. Future work will focus on multimodal integration with whole-slide images, diversification-aware retrieval (e.g., MMR), critique-agent verification pipelines, and tighter integration with enterprise clinical data sources. We anticipate that such advances will further accelerate translational research, improve diagnostic consistency, and help unlock the full clinical value embedded within pathology archives.

## Code and data availability

The processed data and underlying code for this study will be made available upon reasonable request to the corresponding author. WSIs and pathology report data for HISTAI are publicly available through huggingface (https://huggingface.co/datasets/histai/HISTAI-metadata). CHTN data access is subject to institutional data use agreements. Mistral-7B-Instruct model weights are available for research use under Apache 2.0 license and are available at https://huggingface.co/mistralai/Mistral-7B-Instruct-v0.3.

## Author Contributions

A.R.A. conceptualized and designed the study; developed the overall framework and architectural design; implemented the methodology; conducted experiments; and wrote the manuscript. S.W.M. implemented the experiments and computed the results under the guidance of ARA; and edited the manuscript. A.L. and L.G. performed clinical review of all use cases requiring human evaluation; and reviewed and revised the manuscript. A.P. provided clinical insight and access to data; approved the study design; and reviewed and revised the manuscript. R.S. and W.C. provided clinical insight; approved the study design; evaluated use cases as an expert attending pathologist; and reviewed and revised the manuscript. M.K.K.N. conceptualized and designed the study; supervised the research; validated the methodology and results; provided funding; and edited and revised the manuscript.

## Competing Interests

The authors declare no competing interests.

## Acknowledgments

We thank the patients who contributed samples to the CHTN and HISTAI cohorts. We acknowledge HISTAI and CHTN for providing public access to their data, and also Mistral AI for releasing weights for Mistral-7b-Instruct model. We also gratefully acknowledge the Ohio Supercomputer Center for providing high-performance computing resources as part of its contract with The Ohio State University College of Medicine. We also thank the Department of Pathology and the Comprehensive Cancer Center at The Ohio State University for their support.

## Funding

The project described was supported in part by R01 CA276301 (PIs: Niazi and Chen) from the National Cancer Institute, Pelotonia under IRP CC13702 (PIs: Niazi, Vilgelm, and Roy), The Ohio State University Department of Pathology and Comprehensive Cancer Center. The content is solely the responsibility of the authors and does not necessarily represent the official views of the National Cancer Institute or National Institutes of Health or The Ohio State University.

## Ethics Approval and Consent to Participate

This study involved secondary analysis of retrospective, fully de-identified clinical and histopathology data obtained from existing institutional and public repositories. In accordance with applicable regulations, the use of de-identified data does not constitute human subjects' research. Therefore, Institutional Review Board (IRB) approval was not required, and informed consent to participate was waived.

## Supplementary Materials

***Supplementary Table S1:*** *Recall of target report for keyword-based query vs natural language-based query.*

| Retrieval Method | Recall @ 1 | Recall @ 3 | Recall @ 10 |
|---|---|---|---|
| **Keyword Query (traditional)** | 0.1875 | 0.3125 | 0.40625 |
| **Natural Language Query (PathoScribe)** | 0.8125 | 0.90625 | 1.0 |

***Supplementary Table S2:*** *Cosine-similarity diagnostics for target report alignment. Similarities use cosine between the query embedding and (i) the target document embedding and (ii) the best-matching chunk embedding in the target report. Best chunk score (top-100) is the maximum similarity among retrieved top-k chunks that belong to the target report.*

| Model | Pool | $cos(q, d_{gold})$ | $Oracle\ max\ cos(q, c_{gold})$ | Best chunk score (top-100) |
|---|---|---|---|---|
| **PathologyBERT** | CLS | 0.933 [0.930, 0.935] | 0.937 [0.934, 0.940] | 0.954 [0.948, 0.959] |
| **PathologyBERT** | MEAN | 0.599 [0.579, 0.619] | 0.660 [0.636, 0.682] | 0.744 [0.729, 0.760] |
| **e5-large-v2** | CLS | 0.834 [0.829, 0.839] | 0.848 [0.841, 0.855] | 0.855 [0.850, 0.860] |
| **e5-large-v2** | MEAN | 0.858 [0.855, 0.861] | 0.864 [0.859, 0.868] | 0.869 [0.865, 0.873] |
| **bge-m3** | CLS | 0.771 [0.761, 0.780] | 0.759 [0.744, 0.774] | 0.787 [0.775, 0.799] |
| **bge-m3** | MEAN | 0.847 [0.841, 0.852] | 0.880 [0.872, 0.888] | 0.905 [0.899, 0.911] |
| **S-PubMedBert-MS-MARCO** | CLS | 0.984 [0.983, 0.985] | 0.983 [0.981, 0.984] | 0.985 [0.984, 0.986] |
| **S-PubMedBert-MS-MARCO** | MEAN | 0.973 [0.971, 0.974] | 0.970 [0.967, 0.973] | 0.975 [0.973, 0.977] |

***Supplementary Table S3:*** *Doc-level retrieval quality (FAISS dense-doc index; K = 200). Bold indicates the best (highest) mean across all model and pooling configurations within each column. Metrics are reported through R@20.*

| Model | Pool | MRR@200 | nDCG@200 | R@1 | R@5 | R@10 | R@20 |
|---|---|---|---|---|---|---|---|
| **PathologyBERT** | CLS | 0.001 [0.000, 0.003] | 0.013 [0.005, 0.022] | 0.000 [0.000, 0.000] | 0.000 [0.000, 0.000] | 0.000 [0.000, 0.000] | 0.009 [0.000, 0.028] |
| **PathologyBERT** | MEAN | 0.033 [0.011, 0.063] | 0.083 [0.055, 0.116] | 0.019 [0.000, 0.047] | 0.038 [0.009, 0.075] | 0.066 [0.019, 0.113] | 0.104 [0.047, 0.160] |
| **e5-large-v2** | CLS | 0.503 [0.419, 0.584] | 0.585 [0.512, 0.655] | 0.396 [0.302, 0.491] | 0.613 [0.519, 0.708] | 0.708 [0.613, 0.792] | 0.745 [0.660, 0.821] |
| **e5-large-v2** | MEAN | 0.529 [0.447, 0.610] | 0.612 [0.541, 0.682] | 0.425 [0.330, 0.519] | 0.613 [0.519, 0.708] | 0.717 [0.632, 0.802] | 0.783 [0.698, 0.858] |
| **bge-m3** | CLS | 0.503 [0.419, 0.584] | 0.580 [0.505, 0.652] | 0.396 [0.302, 0.491] | 0.604 [0.509, 0.698] | 0.698 [0.613, 0.783] | 0.745 [0.660, 0.830] |
| **bge-m3** | MEAN | 0.259 [0.190, 0.333] | 0.351 [0.285, 0.419] | 0.189 [0.113, 0.264] | 0.330 [0.245, 0.425] | 0.415 [0.321, 0.509] | 0.491 [0.396, 0.585] |
| **S-PubMedBert-MS-MARCO** | CLS | 0.413 [0.333, 0.494] | 0.509 [0.438, 0.579] | 0.311 [0.226, 0.406] | 0.538 [0.443, 0.632] | 0.604 [0.509, 0.698] | 0.642 [0.547, 0.726] |
| **S-PubMedBert-MS-MARCO** | MEAN | 0.387 [0.311, 0.465] | 0.490 [0.423, 0.557] | 0.274 [0.189, 0.358] | 0.519 [0.425, 0.613] | 0.594 [0.500, 0.689] | 0.698 [0.604, 0.783] |

***Supplementary Table S4:*** *Chunk-level retrieval quality (FAISS dense-chunk index; K = 200). Bold indicates the best (highest) mean across all model and pooling configurations within each column. Metrics are reported through R@20.*

| Model | Pool | MRR@200 | nDCG@200 | R@1 | R@5 | R@10 | R@20 |
|---|---|---|---|---|---|---|---|
| **PathologyBERT** | CLS | 0.007 [0.001, 0.018] | 0.020 [0.007, 0.037] | 0.000 [0.000, 0.000] | 0.009 [0.000, 0.028] | 0.019 [0.000, 0.047] | 0.028 [0.000, 0.066] |
| **PathologyBERT** | MEAN | 0.115 [0.066, 0.170] | 0.178 [0.128, 0.234] | 0.075 [0.028, 0.132] | 0.132 [0.075, 0.198] | 0.208 [0.132, 0.292] | 0.255 [0.170, 0.340] |
| **e5-large-v2** | CLS | 0.554 [0.468, 0.638] | 0.619 [0.542, 0.693] | 0.481 [0.387, 0.575] | 0.623 [0.528, 0.717] | 0.679 [0.585, 0.764] | 0.745 [0.660, 0.830] |
| **e5-large-v2** | MEAN | 0.539 [0.451, 0.626] | 0.599 [0.520, 0.677] | 0.491 [0.396, 0.585] | 0.575 [0.481, 0.670] | 0.642 [0.547, 0.736] | 0.745 [0.660, 0.821] |
| **bge-m3** | CLS | 0.345 [0.265, 0.428] | 0.421 [0.345, 0.498] | 0.292 [0.208, 0.377] | 0.415 [0.321, 0.509] | 0.462 [0.368, 0.557] | 0.519 [0.425, 0.613] |
| **bge-m3** | MEAN | 0.310 [0.229, 0.393] | 0.362 [0.284, 0.441] | 0.274 [0.189, 0.358] | 0.349 [0.264, 0.443] | 0.377 [0.283, 0.472] | 0.425 [0.330, 0.519] |
| **S-PubMedBert-MS-MARCO** | CLS | 0.410 [0.327, 0.492] | 0.488 [0.412, 0.563] | 0.330 [0.245, 0.415] | 0.472 [0.377, 0.566] | 0.547 [0.453, 0.642] | 0.613 [0.519, 0.708] |
| **S-PubMedBert-MS-MARCO** | MEAN | 0.414 [0.329, 0.499] | 0.488 [0.411, 0.565] | 0.349 [0.255, 0.443] | 0.472 [0.377, 0.566] | 0.538 [0.443, 0.632] | 0.623 [0.528, 0.717] |

***Supplementary Table S5:*** *IHC panel recommendation: overall and organ-stratified performance (PathoScribe vs LLM-only). All values are percentages (%). Bracketed intervals are paired bootstrap 95% CIs (2,000 iterations). BR@5: budgeted recall within the top-5 markers. Organ-stratified values are macro means within each organ.*

***A.*** *System-level performance (n = 270)*

| Metric | PathoScribe | LLM-only | Δ (PathoScribe-LLM) |
|---|---|---|---|
| **Ranking / early-hit behavior** | | | |
| **Hit@1** | 57.4 [51.5, 63.3] | 11.9 [8.1, 15.9] | +45.6 [38.9, 52.2] |
| **Hit@3** | 76.7 [71.5, 81.9] | 44.8 [38.9, 50.4] | +31.9 [25.2, 38.5] |
| **Hit@5** | 83.7 [79.3, 88.1] | 64.4 [58.5, 70.0] | +19.3 [13.3, 25.6] |
| **BR@5** | 60.8 [56.4, 65.2] | 36.9 [32.7, 41.3] | +23.9 [18.3, 29.5] |
| **Set agreement (marker overlap)** | | | |
| **Precision** | 40.9 [37.5, 44.3] | 23.3 [20.8, 25.9] | +17.6 [14.5, 20.9] |
| **Recall** | 65.0 [60.6, 69.5] | 42.6 [38.3, 47.1] | +22.3 [17.2, 27.5] |
| **Macro F1** | 46.1 [42.8, 49.3] | 27.7 [25.0, 30.5] | +18.3 [14.9, 21.8] |
| **Jaccard** | 34.2 [31.3, 37.2] | 18.4 [16.3, 20.5] | +15.8 [12.9, 18.9] |

***B.*** *Organ-stratified macro performance (95% CIs)*

| Organ | n | F1 | Hit@1 | Hit@3 | Hit@5 | BR@5 |
|---|---|---|---|---|---|---|
| **LLM-only [95% CI] → PathoScribe [95% CI]** | | | | | | |
| **Breast** | 49 | 24.9 [19.6, 30.3] → 49.9 [44.1, 55.9] | 2.0 [0.0, 6.1] → 81.6 [71.4, 91.8] | 26.5 [14.3, 38.8] → 95.9 [89.8, 100.0] | 61.2 [46.9, 73.5] → 95.9 [89.8, 100.0] | 29.9 [21.0, 39.2] → 77.9 [69.5, 85.5] |
| **Colon** | 57 | 22.0 [16.3, 28.1] → 45.7 [37.2, 54.1] | 15.8 [7.0, 26.3] → 50.9 [36.8, 63.2] | 36.8 [24.6, 49.1] → 70.2 [57.9, 80.7] | 50.9 [38.6, 63.2] → 75.4 [63.2, 86.0] | 30.8 [22.0, 40.6] → 56.9 [46.5, 66.9] |
| **Kidney** | 36 | 27.4 [20.5, 34.7] → 51.0 [41.6, 59.9] | 5.6 [0.0, 13.9] → 55.6 [38.9, 72.2] | 58.3 [41.7, 75.0] → 77.8 [63.9, 88.9] | 69.4 [55.6, 83.3] → 86.1 [75.0, 97.2] | 28.6 [21.1, 36.2] → 58.1 [47.4, 68.5] |
| **Liver** | 16 | 33.1 [20.3, 46.7] → 40.6 [25.4, 55.9] | 12.5 [0.0, 31.3] → 50.0 [25.0, 75.0] | 37.5 [12.5, 62.5] → 62.5 [37.5, 87.5] | 68.8 [43.8, 87.5] → 68.8 [43.8, 87.5] | 49.6 [29.7, 69.3] → 57.9 [36.8, 78.4] |
| **Lung** | 77 | 31.7 [26.5, 36.9] → 41.9 [36.6, 47.7] | 14.3 [6.5, 23.4] → 51.9 [40.3, 62.4] | 57.1 [46.8, 68.8] → 71.4 [61.0, 81.8] | 70.1 [59.7, 80.5] → 84.4 [76.6, 92.2] | 43.6 [35.2, 52.1] → 54.2 [46.6, 62.1] |
| **Skin** | 28 | 28.6 [20.2, 37.5] → 48.8 [38.8, 58.5] | 17.9 [3.6, 32.1] → 53.6 [35.7, 71.4] | 42.9 [25.0, 60.7] → 78.6 [60.7, 92.9] | 67.9 [50.0, 85.7] → 85.7 [71.4, 96.4] | 44.6 [29.8, 59.5] → 63.3 [49.2, 76.5] |

***Supplementary Table S6:*** *Reference-based overlap and semantic alignment across report renderings. Values are mean [95% CI]. Higher is better for all metrics. ROUGE and BLEU are computed against available reference rewrites (clinical, oncologist, tumor-board: n=50; patient-friendly: n=13; CAP-style synoptic: n=6). Embedding cosine similarity and PathBERTScore are computed on the full cohort (n=50). 95% CIs for continuous metrics are bootstrap CIs.*

| Setting | ROUGE-1 F1 | ROUGE-2 F1 | ROUGE-L F1 | BLEU-4 | Emb. cos. Sim. | PathBERTScore |
|---|---|---|---|---|---|---|
| **Concise clinical summary** | 0.614 [0.585, 0.644] | 0.410 [0.377, 0.446] | 0.495 [0.461, 0.529] | 0.308 [0.271, 0.346] | 0.819 [0.796, 0.840] | 0.718 [0.701, 0.735] |
| **Oncologist-facing report** | 0.635 [0.593, 0.678] | 0.505 [0.458, 0.551] | 0.581 [0.536, 0.626] | 0.310 [0.256, 0.365] | 0.838 [0.808, 0.868] | 0.780 [0.756, 0.802] |
| **Tumor board-facing report** | 0.655 [0.621, 0.685] | 0.468 [0.437, 0.500] | 0.522 [0.491, 0.552] | 0.362 [0.329, 0.396] | 0.839 [0.803, 0.868] | 0.780 [0.759, 0.799] |
| **CAP-style synoptic** | 0.697 [0.519, 0.841] | 0.608 [0.432, 0.757] | 0.585 [0.414, 0.735] | 0.519 [0.333, 0.681] | 0.987 [0.974, 0.997] | 0.976 [0.957, 0.993] |
| **Patient-friendly report** | 0.379 [0.334, 0.414] | 0.186 [0.164, 0.202] | 0.269 [0.237, 0.294] | 0.121 [0.102, 0.138] | 0.952 [0.927, 0.975] | 0.927 [0.891, 0.961] |

***Supplementary Table S7:*** *Source-grounded retention, faithfulness guardrails, and readability. Values are mean [95% CI]. Higher is better for CSR fact coverage recall, numeric precision, auditor-supported rate, and Flesch Reading Ease; lower is better for FK grade. Metrics are computed on the full cohort (n=50). 95% CIs are bootstrap CIs for continuous metrics; Wilson intervals for proportions (numeric precision; auditor-supported rate).*

| Setting | CSR fact coverage recall | Numeric precision | Auditor-supported | FK grade | Reading Ease |
|---|---|---|---|---|---|
| **Concise clinical summary** | 0.602 [0.560, 0.644] | 1.000 [0.929, 1.000] | 0.740 [0.604, 0.841] | 14.27 [12.97, 15.76] | 20.01 [14.60, 24.94] |
| **Oncologist-facing report** | 0.699 [0.664, 0.733] | 1.000 [0.929, 1.000] | 0.920 [0.812, 0.968] | 12.60 [11.40, 13.82] | 25.46 [19.79, 31.09] |
| **Tumor board-facing report** | 0.650 [0.605, 0.693] | 1.000 [0.929, 1.000] | 0.700 [0.562, 0.809] | 12.73 [11.78, 13.77] | 22.32 [17.58, 26.51] |
| **CAP-style synoptic** | 0.659 [0.617, 0.701] | 1.000 [0.929, 1.000] | 0.660 [0.522, 0.776] | 14.50 [12.83, 16.42] | 11.19 [5.40, 16.62] |
| **Patient-friendly report** | 0.455 [0.421, 0.489] | 1.000 [0.929, 1.000] | 0.820 [0.692, 0.902] | 5.86 [5.54, 6.17] | 69.40 [67.08, 71.58] |

***Supplementary Table S8:*** *Number of local LLM calls and time taken to generate each cohort with average and median over all 10 generations.*

| Cohort Number | Local LLM Calls | Time Taken (seconds) | Time Taken (Minutes) |
|---|---|---|---|
| 1 | 262 | 60.723 | 1.012 |
| 2 | 1426 | 315.11 | 5.252 |
| 3 | 1555 | 356.377 | 5.94 |
| 4 | 2395 | 534.23 | 8.904 |
| 5 | 772 | 174.619 | 2.91 |
| 6 | 1278 | 283.121 | 4.719 |
| 7 | 3608 | 826.199 | 13.77 |
| 8 | 1414 | 304.001 | 5.067 |
| 9 | 2332 | 513.73 | 8.562 |
| 10 | 9446 | 2207.139 | 36.786 |
| **Average** | **2448.8** | **557.5249** | **9.2922** |
| **Median** | **1555** | **356.377** | **5.94** |

## S1 Supplementary Methods

### S1.1 OCR Rendering and OCR Configuration

CHTN PDFs were rendered page-wise to grayscale images at 350 dpi prior to OCR. OCR was performed with Tesseract using English language data, OEM set to 1 and PSM set to 6, with a per-page timeout of 120 seconds. Page text was optionally written to per-page TXT files and was always merged into a single document TXT file with explicit page boundary markers retained for traceability.

### S1.2 Post-OCR Normalization and Section Parsing

Post-OCR normalization includes removal of explicit page markers, repair of hyphenation across line breaks, collapsing of broken lines while preserving paragraph structure, and whitespace normalization. Section parsing uses regular expressions over common pathology report headings, including FINAL DIAGNOSIS, DIAGNOSIS, MICROSCOPIC DESCRIPTION, GROSS DESCRIPTION, COMMENT, and IMMUNOHISTOCHEMISTRY headings. Sections are retained as labeled blocks and are used to constrain chunk boundaries.

### S1.3 Chunk Construction and JSONL Artifacts

Section text is split into sentences using a punctuation-based boundary heuristic and then assembled into greedy chunks that respect minimum and maximum token thresholds. Chunks do not cross section boundaries. The pipeline emits (i) a document-level JSONL file containing document identifiers and provenance paths and (ii) a chunk-level JSONL file containing chunk identifiers, document identifiers, section labels, and chunk text. Chunks include a short extractive summary computed as the first three sentences of the chunk.

### S1.4 Optional Metadata and Summary Generation

An optional step extracts document-level metadata and produces short chunk summaries using an on-premises LLM served via an OpenAI-compatible API. Metadata fields include title, accession number, date, specimen site, diagnosis, IHC markers, keywords, and a brief summary. These fields are used for indexing and user-facing display and do not alter dense retrieval scoring.

### S1.5 Hybrid Weights and Default Settings

Hybrid retrieval uses the convex combination in hybrid-score with default weights $\alpha_{doc} = 0.5, \alpha_{chunk} = 0.3, and\ \alpha_{bm25} = 0.2,$ satisfying $\alpha_{doc} + \alpha_{chunk} + \alpha_{bm25} = 1.$

### S1.6 Cost Analysis of NVIDIA A100 40GB GPU On-Demand and On-Prem

To estimate the per-hour cost of an NVIDIA A100 40GB GPU on-demand, we consulted the Amazon AWS EC2 pricing page on 03/04/2026 at 1:31 P.M. The relevant instance, p4d.24xlarge, includes 8 A100 40GB GPUs and costs approximately $21.96 per hour in the US East (Ohio) region, translating to roughly $2.74 per GPU per hour. The cost of new hardware was approximated using reports from Northflank and Jarvis Labs (2025), which estimate the price of an A100 40GB PCIe GPU at $8,000–12,000.

*(Webpages accessed: Amazon EC2 P4 Instances: https://aws.amazon.com/ec2/instance-types/p4/; Amazon EC2 On-Demand Pricing: https://aws.amazon.com/ec2/pricing/on-demand/; Northflank Article: https://northflank.com/blog/nvidia-a100-gpu-cost; Jarvis Labs Article: https://jarvislabs.ai/ai-faqs/nvidia-a100-gpu-price)*